\documentclass{article}

\usepackage[preprint]{neurips_2024}

\usepackage[utf8]{inputenc} 
\usepackage[T1]{fontenc}    
\usepackage{hyperref}       
\usepackage{url}            
\usepackage{booktabs}       
\usepackage{amsfonts}       
\usepackage{nicefrac}       
\usepackage{microtype}      
\usepackage{xcolor}         

\usepackage{amsmath}
\usepackage{graphicx}
\usepackage{caption}
\usepackage{subcaption}

\usepackage{algorithm}
\usepackage{algpseudocode}

\newif\ifcomments
\commentstrue 
\ifcomments
    \newcommand{\lianmin}[1]{{\color{blue}{\bf\sf [Lianmin: #1]}}}
    \newcommand{\ying}[1]{{\color{blue}{\bf\sf [Ying: #1]}}}
    \newcommand{\ion}[1]{{\color{blue}{\bf\sf [Ion: #1]}}}
    \newcommand{\joey}[1]{{\color{blue}{\bf\sf [Joey: #1]}}}
    \newcommand{\todo}[1]{{\color{orange}{\bf\sf [TODO: #1]}}}
    \newcommand{\fixme}[1]{{\color{orange}{#1}}}
\else
    \newcommand{\lianmin}[1]{}
    \newcommand{\ying}[1]{}
    \newcommand{\ion}[1]{}
    \newcommand{\joey}[1]{}
    \newcommand{\todo}[1]{}
    \newcommand{\fixme}[1]{}
\fi

\algnewcommand{\LineComment}[1]{\State \textcolor{blue}{\# #1}}

\title{Post-Training Sparse Attention with Double Sparsity}

\author{%
  Shuo Yang  $^{13}$ \quad
  Ying Sheng$^{2}$ \quad
  Joseph E. Gonzalez$^{1}$ \quad
  Ion Stoica${^1}$ \quad
  Lianmin Zheng$^{1}$ \\
  $^1$UC Berkeley \quad
  $^2$Stanford \quad
  $^3$Shanghai Jiao Tong University
}

\begin{document}

\maketitle

\begin{abstract}

The inference process for large language models is slow and memory-intensive, with one of the most critical bottlenecks being excessive Key-Value (KV) cache accesses. This paper introduces "Double Sparsity," a novel post-training sparse attention technique designed to alleviate this bottleneck by reducing KV cache access. Double Sparsity combines token sparsity, which focuses on utilizing only the important tokens for computing self-attention, with channel sparsity, an approach that uses important feature channels for identifying important tokens. Our key insight is that the pattern of channel sparsity is relatively static, allowing us to use offline calibration to make it efficient at runtime, thereby enabling accurate and efficient identification of important tokens. Moreover, this method can be combined with offloading to achieve significant memory usage reduction.
Experimental results demonstrate that Double Sparsity can achieve \(\frac{1}{16}\) token and channel sparsity with minimal impact on accuracy across various tasks, including wiki-2 perplexity, key-value retrieval, and long context benchmarks with models including Llama-2-7B, Llama-2-70B, and Mixtral-8x7B.
It brings up to a 14.1$\times$ acceleration in attention operations and a 1.9$\times$ improvement in end-to-end inference on GPUs. With offloading, it achieves a decoding speed acceleration of 16.3$\times$ compared to state-of-the-art solutions at a sequence length of 256K.
Our code is publicly available at \url{https://github.com/andy-yang-1/DoubleSparse}.

\end{abstract}

\section{Introduction}

Large Language Models (LLMs) have significantly advanced machine learning capabilities, enabling a wide range of applications from natural language processing to complex problem-solving tasks~\citep{achiam2023gpt,touvron2023llama,team2023gemini}. However, their inference remains costly and slow due to token-by-token decoding. This decoding process exhibits low arithmetic intensity, making it largely memory-bound. During decoding, access to two types of memory is required: model weights and the Key-Value (KV) cache in the self-attention layers~\citep{vaswani2017attention}. Both can be very large and thus become bottlenecks. When the batch size is large or the sequence length is long, the size of the KV cache can easily surpass that of the model weights~\citep{pope2023efficiently}. While extensive research has focused on reducing access to model weights through quantization and sparsification, the reduction of access to the KV cache has received less attention.

In this paper, we explore methods to reduce access to the KV cache during inference, thereby making attention computation more bandwidth-efficient and accelerating its execution. Our focus is on post-training methods that can be directly applied to a pre-trained model to provide wall-clock acceleration without requiring excessive additional training or fine-tuning overhead. Prior work has attempted to leverage quantization~\citep{hooper2024kvquant,liu2024kivi}, compression~\citep{nawrot2024dynamic}, and sparsity~\citep{zhang2024h2o,anagnostidis2024dynamic,ge2024model,ribar2023sparq} to achieve these goals.
Among them, sparsity holds significant potential if a high sparsity ratio can be achieved. The intuition of sparsification is that not every token is equally important for decoding the next token. Therefore, during the decoding process, we can rely on a small subset of important tokens to compute the self-attention, achieving nearly the same results.
While the approach of sparse attention seems intuitive, previous research has struggled to find a post-training sparse attention method that maintains high accuracy while being runtime-efficient.

The primary challenge in post-training sparse attention lies in accurately and efficiently identifying important tokens. A naive approach entails calculating the entire attention weight matrices and then sorting the tokens based on the accumulated attention weights. Although this method can precisely identify important tokens, it fails to offer a runtime speedup, as it requires computing the full attention weight matrices, which is precisely the step we aim to avoid.
Previous studies have proposed various methods for selecting important tokens; however, these methods either lead to significant accuracy losses or fail to achieve practical wall-clock acceleration. Notably, H2O~\citep{zhang2024h2o} employs a dynamic strategy that maintains a small fixed-size cache of important tokens. Due to its limited size, it must evict many tokens. Since it cannot predict future important tokens, it often inadvertently evicts them, leading to accuracy degradation.
SparQ~\citep{ribar2023sparq}, in contrast, retains all tokens and dynamically selects important ones at each step. Yet, its design falls short of achieving the desired speedup and incurs considerable additional memory overhead. In summary, designing an efficient method capable of accurately identifying important tokens remains a significant challenge.

We propose ``Double Sparsity,'' a method that leverages both token sparsity and channel sparsity to achieve accurate and efficient post-training sparse attention. Token sparsity refers to the sparse attention method mentioned above~\citep{zhang2024h2o}, which uses only important tokens to compute self-attention. Channel sparsity, a new method we introduced, selects important tokens at runtime using significant feature channels. Our key insight is that while token sparsity is highly dynamic, channel sparsity exhibits relatively static behavior, enabling us to identify and select important channels through offline calibration. This static channel sparsity thus provides an efficient means to achieve dynamic token sparsity at runtime. Building on this concept, we carefully explored how to select important channels based on statistics from offline calibrations.
Furthermore, once we can quickly identify important tokens for the current layer, we extend this process by predicting the important tokens of the next layer. We achieve this by utilizing the embedding similarity between adjacent layers. This approach enables us to offload the entire KV cache to host memory and prefetch only the important tokens to GPU memory, significantly reducing GPU memory footprint.

Shown in Figure \ref{fig:Llama-3d-figure}, we demonstrate that Double Sparsity can achieve both an $\frac{1}{16}$ token sparsity and an $\frac{1}{16}$ channel sparsity simultaneously, while incurring only a negligible accuracy loss across a broad array of benchmarks, including language modeling, question answering, and retrieval tasks. The sparsity directly leads to the reduction of memory access and runtime speedup. Double Sparsity accelerates the attention operation by up to $14.1\times$ at a sparsity level of $\frac{1}{16}$ on NVIDIA A10G and A100G GPUs, closely approaching the theoretical acceleration upper bound. It accelerates end-to-end inference for various workloads by up to $1.9\times$. When turning on offloading, it achieves a decoding throughput that is $16.3\times$ higher than the state-of-the-art offloading-based solutions at a sequence length of 256K.
\section{Background}

\label{sec:background}

\subsection{Preliminaries on Self-Attention and Notations}

Attention computation is one of the major bottlenecks in LLM Inference, especially when the sequence length is large~\citep{tay2022efficient}. This is caused by its quadratic computational complexity.
Let $d_h$ denote the head dimension, and $S$ denote the number of tokens.
We use the decoding step as an example to illustrate the self-attention computation.
Each token carries three tensors to embed its information, which are called query, key, and value.
In an attention layer, let $q\in \mathbb{R}^{d_h}$ represents the query tensor for input token, $K\in \mathbb{R}^{S\times d_h}$ represents the key tensor for all tokens, and $V\in \mathbb{R}^{S\times d_h}$ represents the value tensor for all tokens. The attention is obtained through the formula shown below:

\[ y=softmax\left( \frac{q\cdot K^T}{\sqrt{d_h}} \right) \cdot V \]



\subsection{Post-training Sparse Attention}

In this work, we introduce the term "post-training sparse attention," analogous to "post-training quantization." Post-training sparse attention refers to techniques that exploit inherent model sparsity, such as token-level sparsity, to accelerate attention calculations without requiring additional training. In the field of LLMs, many works have utilized post-training sparse attention, including H2O, StreamingLLM~\citep{xiao2024efficient} and SparQ. However, these methods come with significant limitations, presenting serious challenges for post-training sparse attention.


\section{Challenges in Post-Training Sparse Attention}


\label{sec:observation}

In this section, we discuss prior research on post-training sparse attention, identifying the challenges and shortcomings that have prevented these approaches from achieving their full potential.
More related work is included in Section \ref{sec:related_work}.

\subsection{Retrieval Accuracy}

One of the most challenging issues for post-training sparse attention is maintaining retrieval accuracy. For instance, StreamingLLM discards earlier tokens, while H2O selectively drops tokens based on previous attention scores. 
Although discarding tokens can accelerate computations, this exclusion leads to the loss of critical information, potentially compromising the model's retrieval accuracy. As highlighted in \cite{jelassi2024repeat}, this issue is inherent to techniques that rely on discarding tokens, prompting the exploration of sparse attention methods that preserve the complete KV cache.

\subsection{Hardware Friendliness}

Achieving wall-clock speedup poses a greater challenge while maintaining model retrieval accuracy, particularly because some post-training sparse attention techniques are not hardware-friendly. 
For instance, SparQ retains the complete KV cache and computes attention selectively on a subset of the KV cache based on the query. This approach theoretically allows for acceleration while maintaining accuracy.
However, SparQ’s method of selecting channels and tokens results in non-contiguous memory access, causing substantial L1/L2 cache misses and wasting GPU bandwidth with the standard 128-byte memory access. 
Despite being designed to accelerate processing, SparQ achieves only a modest 1.3 times speed increase in attention computations.
Therefore, it is crucial to develop an algorithm that ensures continuous memory access patterns and avoids dynamic selection of channels to accelerate attention while preserving accuracy.

\subsection{Memory Usage}

Methods that preserve the complete KV cache inherently require substantial GPU memory consumption.
To mitigate the heavy memory demand, the FlexGen~\citep{sheng2023flexgen} approach offloads the KV cache of each layer to the GPU only during the computation phase.
However, a significant challenge arises because FlexGen needs to offload the complete KV cache, and the communication overhead can drastically affect overall system performance.
Considering that selected tokens constitute just a small fraction of all tokens, the time taken to offload these specific tokens to the GPU is considerably less than the time required for offloading the entire KV cache as in FlexGen. 
By efficiently managing when and how data is transferred and processed, it's possible to significantly reduce both the time and memory overhead typically associated with maintaining a full KV cache.



To address these challenges, we propose two post-training sparse attention techniques. In Section \ref{sec:double-sparsity}, we introduce Double Sparsity, which accelerates attention by up to 16 $\times$ with minimal additional memory consumption. 
In Section \ref{sec:double-sparsity-offload}, we present Double Sparsity-Offload, which reduces memory usage to 1/16 without increasing latency.
\section{Double Sparsity}

\label{sec:double-sparsity}

\begin{figure}[t]  
  \centering
    \vspace{-1em}
  
  \begin{minipage}{0.42\textwidth}
    \begin{algorithm}[H]
      \caption{Double Sparsity Decode}
      \label{alg:sparse-forward}
      \begin{algorithmic}[1]
        \vspace{5pt}
        \Require $Q \in \mathbb{R}^{d_h}, K \in \mathbb{R}^{S \times d_h},\newline \vspace{5pt} V \in \mathbb{R}^{S \times d_h}, C \in \mathbb{N}^r,\newline \vspace{5pt} K_{label} \in \mathbb{R}^{S \times r}, r = \alpha d_h, k = \beta S$
        \Ensure $y$
        \State $Q_{label} \leftarrow Q_{[C]}$ \vspace{5pt}
        \State $\hat{s} \leftarrow Q_{label} \cdot K_{label}$  \vspace{5pt}
        \State $i \leftarrow \text{argtopk}(\hat{s}, k)$ \vspace{5pt}
        \State $s \leftarrow \text{softmax}\left(\frac{Q \cdot K_{[i,:]}^T}{\sqrt{d_h}}\right)$  \vspace{5pt}
        \State $y \leftarrow s \cdot V_{[i,:]}$  \vspace{5pt}
        \State \textbf{return} $y$ \vspace{5pt}
      \end{algorithmic}
    \end{algorithm}
    \vspace{-10pt}
    \includegraphics[width=\textwidth]{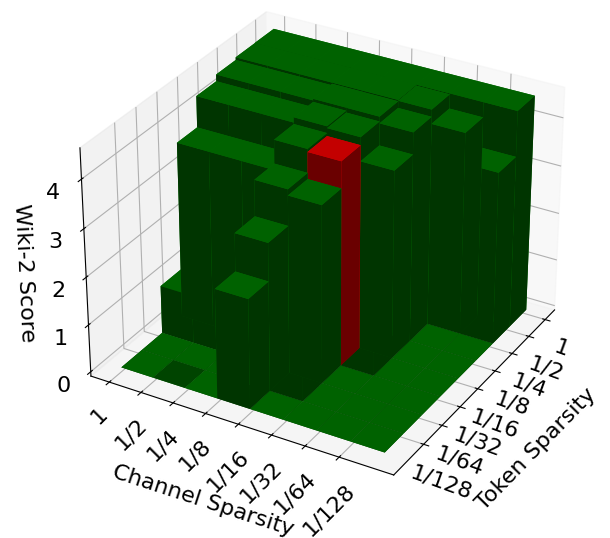}
    \caption{Perplexity of Llama at different token-sparsity and channel-sparsity levels.}
    \label{fig:Llama-3d-figure}
  \end{minipage}%
  \hfill
  \begin{minipage}{0.55\textwidth}
    \includegraphics[width=\textwidth]{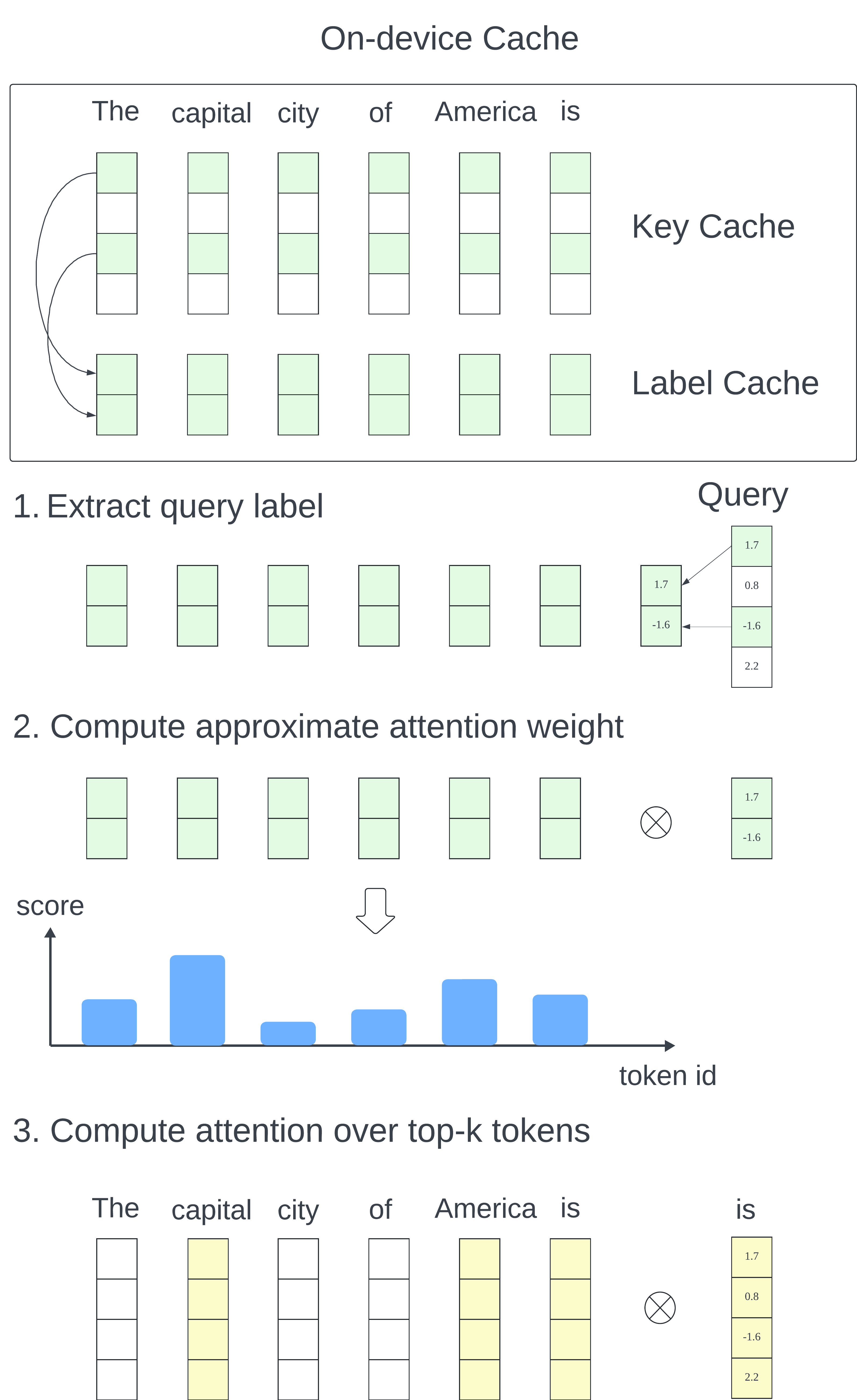}
    \caption{Decoding process of Double Sparsity.}
    \label{fig:decoding-process}
  \end{minipage}
\end{figure}

Based on the insights of Section \ref{sec:observation}, we propose Double Sparsity, a hardware-friendly and bandwidth-efficient post-training sparse attention mechanism. 
This approach overcomes the challenges highlighted in previous post-training sparse attention techniques by ensuring no loss of information, as it maintains the \textbf{entire KV cache}. 
To avoid the cache misses associated with runtime sorting, Double Sparsity utilizes \textbf{offline calibration} to pre-determine outlier channels for each transformer layer. 
A compact \textbf{label cache} is employed to store outlier channel values from the Key cache, optimizing memory access patterns to leverage GPU's preference for contiguous memory access. 
Algorithm \ref{alg:sparse-forward} and Figure \ref{fig:decoding-process} illustrate the decoding process of Double Sparsity.

  

\subsection{Offline Calibration}

Offline calibration is a commonly used technique to identify channel sparsity, particularly effective for pinpointing outlier channels. 
For example, AWQ~\citep{lin2023awq} utilizes offline calibration to identify salient weight channels that significantly impact model performance. 
Inspired by this approach, we employ offline calibration to pre-determine the channels that most influence attention scores. 
Attention computation can be expressed as $A= Q \cdot K^T$ , which can be broken down into $A = \sum_i^{d_h} S_i $ where $S_i =  Q_i * K_i$. Due to channel sparsity, only a few $S_i$ have a significant impact on $A$.
Therefore, by conducting offline calibration on a small validation set, we can efficiently identify these critical channels by computing the $\underset{i}{argmax} \, S_i$.
Figure \ref{fig:offline-calibration} in Appendix \ref{sec:appendix-offline} illustrates the outlier channels identified by AWQ and Double Sparsity.

To validate the efficacy of outlier channels identified through offline calibration, we conducted a comparison in Appendix \ref{sec:appendix-offline} between the outlier channel indices derived from offline calibration and those determined during the online decoding process.
A significant overlap between the two sets underscores the reliability of offline-calibrated outliers.
Figure \ref{fig:calibration-overlap} illustrates this relationship. 
An observation from the comparison is that when the ratio surpasses 0.25, the overlap reaches 0.95.



\subsection{Forwarding with Label Cache}


After identifying the outlier channel indices, it becomes crucial to access them efficiently.
Reading these channels directly from the Key cache can lead to non-contiguous memory accesses, which significantly underutilized the bandwidth.
To alleviate non-contiguous memory accesses, we leverage a label cache to store pre-determined heavy channel values.
This label cache allows for continuous memory access when computing approximate attention, avoiding the need to retrieve non-contiguous segments from the Key cache. 
During the prefilling stage, all heavy channel values from the Key cache are stored in the label cache; in the decoding phase, only the heavy channel values of new tokens are added.
Since approximate attention is not sensitive to precision, we can store the label cache in 4-bit.
This approach enables us to maintain a label cache that is only 1/16 the size of the K cache, facilitating contiguous memory access and significantly improving the hit rate of L1/L2 caches, thereby optimizing inference speed and efficiency.
In Appendix \ref{sec:no-label-cache}, an ablation study was conducted to evaluate the impact of label caches. The results demonstrated that a label cache accelerates decoding speeds by 2 to 4 times compared to configurations without a label cache.






\section{Reducing GPU Memory Usage with Double Sparsity-Offload}

\label{sec:double-sparsity-offload}

Building upon Double Sparsity, we propose the Double Sparsity-Offload technique to further reduce the GPU memory overhead in large language models.
This approach significantly diminishes the memory requirement to 1/16 of the original KV caches. By optimizing memory usage, Double Sparsity-Offload enables more efficient decoding, especially with limited GPU memory resources.

\subsection{Prefetching Tokens with Double Buffer}

The Double Sparsity-Offload algorithm introduces a double buffer prefetching system during for decoding process.
The complete KV cache is stored on the CPU, while the GPU maintains only the label cache and a double buffer.
During the decoding process, each layer processes its embeddings through the next layer’s query projection to generate an approximate query for the subsequent layer. This approximate query is then used to compute the next layer's approximate attention. While the current layer’s attention and feed-forward network computations are being performed, the tokens corresponding to the approximate attention results for the next layer are offloaded to the GPU. This use of double buffering allows for a smooth and efficient overlap of computation and memory transfer.

\subsection{Empirical Analysis: Embedding Similarity Between Layers}

The feasibility of the Double Sparsity-Offload algorithm is based on the high degree of similarity between embeddings across consecutive layers. 
To empirically validate this assumption, we conducted an analysis using the Pile validation dataset, applied to the Llama-2-7B model. We measured the cosine similarity of embeddings between every two consecutive layers throughout the model. The results show that apart from the first two layers, the second and third layers, and the very last layers (30 and 31), all other layer pairs exhibited a cosine similarity exceeding 90\%, with the majority of layers showing similarities above 95\%. These high similarity scores support the viability of utilizing prior layer embeddings to predict queries for subsequent layers in Double Sparsity-Offload. 


\subsection{Complexity Analysis}

To understand the potential speedup of Double Sparsity, we need to analyze its Cache IO-Complexity since attention mechanisms are bandwidth-bounded.
Double Sparsity can be simplified into two steps: calculating approximate attention and computing attention over $k$ tokens. 
Memory-wise, the total access comprises $O(d)$ bytes for $Q$, $O(S\times r)$ for the label cache, $O(2\times k \times d)$ for the KV cache, leading to a total of $O(S\times r + 2\times k \times d) = O(\alpha\times S\times d + 2\times \beta \times S \times d)$.
Given that the approximate attention phase of Double Sparsity does not involve softmax operations, it allows for high parallelism compared to the following step.
Therefore, the overall IO complexity of Double Sparsity primarily depends on the latter step, which can be approximated as $O(2\times \beta \times S \times d)$.
This analysis reveals that Double Sparsity’s time complexity is linearly dependent on $\beta$, and the extra memory overhead is linearly proportional to $\alpha$. 
Table \ref{tab:sparsity_work_comparison} summarizes all the sparsity works discussed, specifying their overhead, complexity, and speedup.

\begin{table}[t]
\centering
\vspace{-1.4em}
\caption{Comparison of sparsity-related techniques. `SparQ (1xK)' denotes single-dimension storage of the Key cache, while `SparQ (2xK)' refers to dual-dimension storage of the Key cache.}
\label{tab:sparsity_work_comparison}
\begin{tabular}{lcccc}
\toprule
Method             & On-device Cache Size        & Cache IO-Complexity   & Min $\beta$ & Speedup \\
\midrule
H2O               & $S \times \beta$  & $S \times \beta$ & 1/5        & Yes     \\
SparQ (1xK)         & $S$                 & $S \times \beta$ & 1/8        & No      \\
SparQ (2xK)         & $S \times 1.5$    & $S \times \beta$ & 1/8        & Yes     \\
AWQ              & $S$                 & $S$                & 1          & Yes     \\
Double Sparsity  & $S \times (1 + \frac{\alpha}{2})$ & $S \times \beta$ & 1/16     & Yes     \\
Double Sparsity-Offload  & $S \times \frac{\alpha}{2}$ & $S \times \beta$ & 1/16     & Yes     \\
\bottomrule
\end{tabular}
\end{table}

\section{Experiment}

\label{sec:exp}

In Section \ref{sec:accuracy-eval}, we demonstrate that both Double Sparsity and Double Sparsity-Offload maintain robust performance with a sparsity setting of 1/16 across various benchmarks, including Wiki-2 perplexity~\citep{merity2016pointer}, MultifieldQA~\citep{bai2023longbench}, GovReport~\citep{huang2021efficient}, TriviaQA~\citep{2017arXivtriviaqa}, and MMLU~\citep{hendryckstest2021}. 
In key-value retrieval tasks, Double Sparsity significantly outperforms other post-training sparse attention techniques. 
In Section \ref{sec:speedup-eval}, we compare Double Sparsity against state-of-the-art attention and end-to-end implementations. 
Results show that Double Sparsity achieves up to a 16-fold acceleration in attention mechanisms and up to a twofold increase in overall end-to-end processing speed. Additionally, Double Sparsity-Offload achieves a 16-fold acceleration compared to FlexGen Offload.

\subsection{Accuracy Evaluation}

\label{sec:accuracy-eval}

\subsubsection{Wiki-2 Perplexity}

Wiki-2 perplexity is a benchmark derived from Wikipedia articles, offering a comprehensive test with its broad vocabulary and authentic text features. 
A lower score indicates better model performance. 
Table \ref{tab:model-sparsity-level} illustrates the changes in perplexity across different sparsity levels for each model.

\begin{table}[t]
\centering
\vspace{-1em}
\caption{Perplexity of models at various sparsity levels. Note the minimal changes in perplexity from sparsity levels 1 to 1/16, with a significant performance gap emerging between levels 1/16 and 1/32.}
\label{tab:model-sparsity-level}
\begin{tabular}{lcccccc}
\toprule
\multicolumn{1}{c}{} & \multicolumn{6}{c}{Sparsity Level} \\
\cmidrule{2-7}
Model           & 1    & 1/2  & 1/4    & 1/8    & 1/16    & 1/32 \\
\midrule
Llama-7B        & 5.68 & 5.69 & 5.69     & 5.72     & 5.80     & 7.66 \\
Llama-2-7B      & 5.47 & 5.48 & 5.53 & 5.56 & 5.76  & 12.01 \\
Llama-2-7B (offloading)  & 5.47 & 5.48 & 5.54 & 5.57 & 5.86  & 15.29 \\
Llama-2-7B-chat & 6.94 & 6.94 & 6.96     & 6.92     & 7.14     & 14.93 \\
Llama-2-7B-chat (offloading) & 6.94 & 6.94 & 6.97     & 6.92     & 7.33     & 20.08 \\
Mistral-7B      & 5.25 & 5.25 & 5.26     & 5.27     & 5.37     & 14.55 \\
\bottomrule
\end{tabular}
\end{table}

To demonstrate the model's performance at different sparsity levels and justify our selection of a sparsity level of 1/16, we constructed a 3D bar chart. According to Figure \ref{fig:3d-bar} in Appendix \ref{sec:perplexity-select}, a noticeable shift in perplexity is observed as the sparsity level goes beyond 1/16.

To validate the robustness of Double Sparsity and the effectiveness of the 1/16 sparsity level, we conducted a series of ablation studies across various model configurations and conditions. 
Table \ref{tab:perplexity-ablation} demonstrates the effectiveness of Double Sparsity at a sparsity level of 1/16 across various model sizes, attention mechanisms, and MoE configurations. 


\begin{table}[t]
\centering
\caption{Ablation study on different architectural models with different outlier types at 1/16 sparsity level. Note that GQA models are incompatible with K outlier channel.}
\label{tab:perplexity-ablation}
\begin{tabular}{llccccc}
\toprule
Model          & Architecture  & \multicolumn{1}{c}{Original} & \multicolumn{4}{c}{Double Sparsity}             \\
\cmidrule(lr){4-7}
               &               &                              & random channel & q outlier & k outlier & qk outlier \\
\midrule
Llama-2-7B     & Single/MHA    & 5.47                         & 8.62              & 6.45      & 6.61      & 5.76       \\
Llama-2-7B-chat& Single/MHA    & 6.94                         & 10.1              & 7.8       & 9.44      & 7.14       \\
Mistral-7B     & Single/GQA    & 5.25                         & 6.06              & 5.79      & N/A       & 5.37       \\
Llama-2-70B    & Single/GQA    & 3.32                         & 5.15              & 3.69      & N/A       & 5.17       \\
Mixtral-8x7B   & MoE/GQA       & 3.84                         & N/A              & 3.84      & N/A       & 17.3       \\
\bottomrule
\end{tabular}
\end{table}

\subsubsection{Long Context Benchmarks}

\begin{figure}[t]
\vspace{-1em}
\centering 
\includegraphics[width=\textwidth]{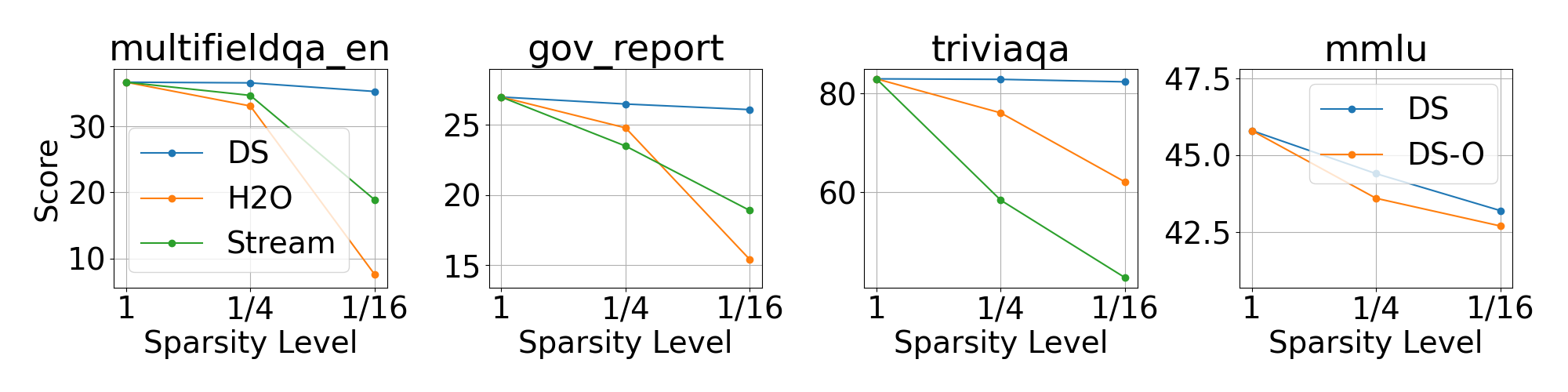} 
\caption{Performance of different techniques across various sparsity levels for long context benchmarks. `DS' and `DS-O' refer to Double Sparsity and Double Sparsity-Offloading. `Stream' refers to Streaming-LLM.} 
\label{fig:longbench} 
\end{figure}

We used Llama-2-7B to evaluate the performance of Double Sparsity across multiple long context benchmarks at various levels of sparsity, comparing its effectiveness with that of StreamingLLM and H2O.
As illustrated in Figure \ref{fig:longbench}, Double Sparsity maintains its performance with nearly no drop in accuracy at a sparsity level of 1/16, outperforming other techniques.

\subsubsection{Key-Value Retrieval}

\begin{figure}[t]
\centering 
\includegraphics[width=\textwidth]{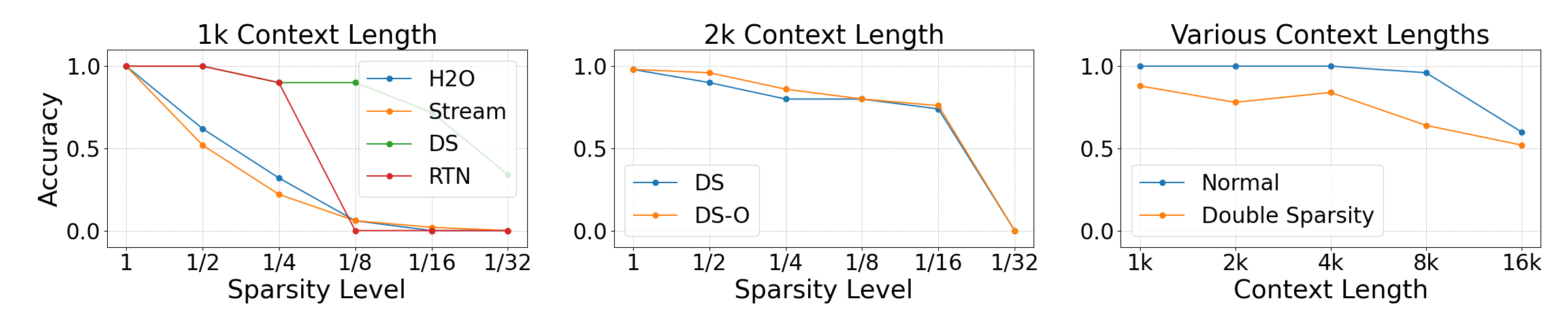} 
\caption{Retrieval accuracy across various sparsity levels and context lengths.
`DS' and `DS-O' refer to Double Sparsity and Double Sparsity-Offloading. `Stream' refers to Streaming-LLM. `RTN' refers to RTN Quantization.} 
\label{fig:kv-retrieval} 
\end{figure}

The key-value retrieval benchmark is designed to assess a model's in-context retrieval capabilities.
Our experiments compared Double Sparsity against other post-training sparsity techniques, including H2O, StreamingLLM, and RTN quantization~\citep{nagel2020down}.
We also tested the performance of Double Sparsity with the Vicuna-7B-16K model to observe how accuracy changes as context length increases.
As shown in Figure \ref{fig:kv-retrieval}, we demonstrate that Double Sparsity significantly surpasses the other techniques in key-value retrieval tasks. 
Notably, Double Sparsity and Double Sparsity-Offload show equivalent performance, highlighting that the offloading mechanism exhibits almost no decay. 

\subsection{Speedup Evaluation}

\label{sec:speedup-eval}


\subsubsection{Setups}

\textbf{Hardware.} Our experiments were conducted on two types of GPUs: the A10G and the A100-SXM. 

\textbf{Implementation.} For the Double Sparsity Attention, we utilized PyTorch to compute approximate attention and select the top-k tokens. The kernel for attention over top-k tokens was designed using OpenAI Triton. 
For end-to-end testing, we replaced the full attention mechanism in gpt-fast~\citep{pytorch2023accelerating} with our Double Sparsity Attention. 
For Double Sparsity-Offload, we implemented asynchronous CPU to GPU memory copying using CUDA streams and DGL~\citep{wang2019dgl}'s gathering kernel.

\textbf{Workload.} We focused on high-workload scenarios to push the limits of Double Sparsity. This included a range of batch sizes from 4 to 32 and sequence lengths from 1024 to 16384. 
For Double Sparsity-Offload, we extended testing to extreme conditions on the A100 GPU, exploring sequence lengths from 64K to 256K.
Given that gpt-fast's KV cache is pre-allocated, the tokens-per-second throughput depends solely on the batch size and sequence length.

\textbf{Baseline.} For attention acceleration evaluations, we use the \textbf{`scaled\_dot\_product\_attention'} as our baseline. This implementation ranks among the fastest attention mechanisms, dynamically allocating computation among the most efficient options including \textbf{FlashAttention-2}~\citep{dao2023flashattention2}, \textbf{Memory-Efficient Attention}~\citep{xFormers2022}, and the top-performing kernels from the PyTorch team. In the end-to-end speed evaluations of Double Sparsity, \textbf{gpt-fast} serves as the baseline, distinguished as the state-of-the-art for Llama models on the A100 GPU. It offers exceptionally low latency and throughput that surpasses that of the huggingface transformers by tenfold. For evaluating Double Sparsity-Offload, we compare it against FlexGen Offloading, which shares the same gpt-fast codebase and memory footprint.

\textbf{Other Settings.} Given Double Sparsity's focus on the attention mechanism, both weights and activations were set to FP16 precision. Furthermore, considering the limitations imposed by Triton kernels on Torch compile options, neither Double Sparsity nor gpt-fast employed the Torch compiler.

\subsubsection{Attention Operator Speedup}

\begin{figure}[t]
\vspace{-10pt}
\centering 
\includegraphics[width=\textwidth]{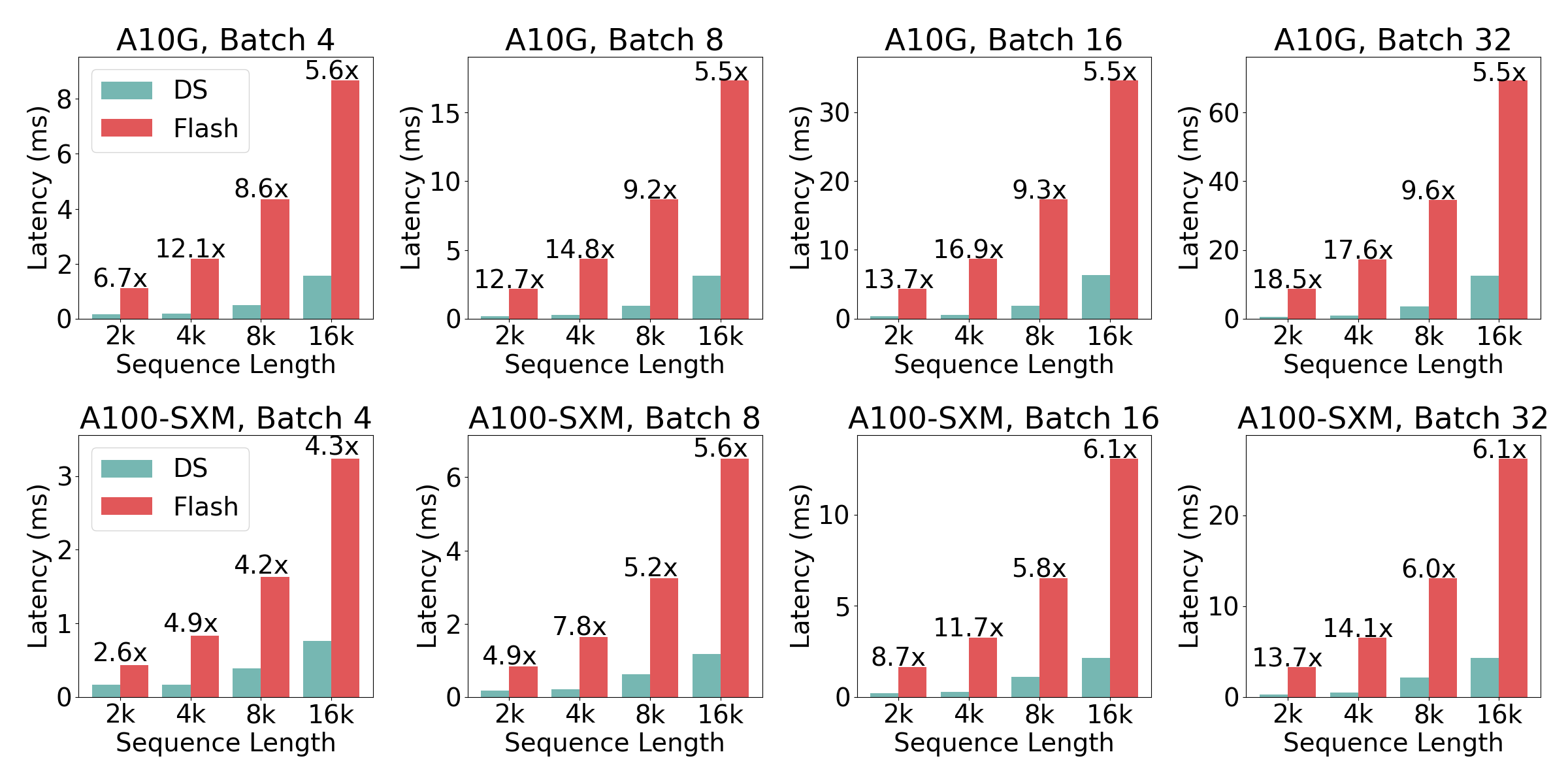} 
\caption{Latency and speedup of Double Sparsity Attention at various batch sizes and sequence lengths. `DS' indicates double sparsity attention. `Flash' indicates the `scaled\_dot\_product\_attention', which is the fastest of FlashAttention-2 and Memory-Efficient Attention.} 
\label{fig:attention-speedup} 
\end{figure}

Figure \ref{fig:attention-speedup} provides a comprehensive view of the latency and speedup of Double Sparsity compared to `scaled\_dot\_product\_attention' across different batch sizes and sequence lengths. 
On the A10G GPU, every case achieves at least a fivefold speedup, with more than half exceeding ninefold. Notably, Double Sparsity achieves a linear speedup at a sequence length of 4096 with large batch sizes. On the A100 GPU, nearly all cases see at least fourfold faster processing, with larger batches reaching up to tenfold speedup. 
The greater speedup for smaller batches on the A10G might be due to the launch time of Triton kernels, which becomes significant when the kernel execution time on the A100 is short.



\subsubsection{End-to-End Inference Speedup}

\begin{figure}[t]
\vspace{-10pt}
\centering 
\includegraphics[width=\textwidth]{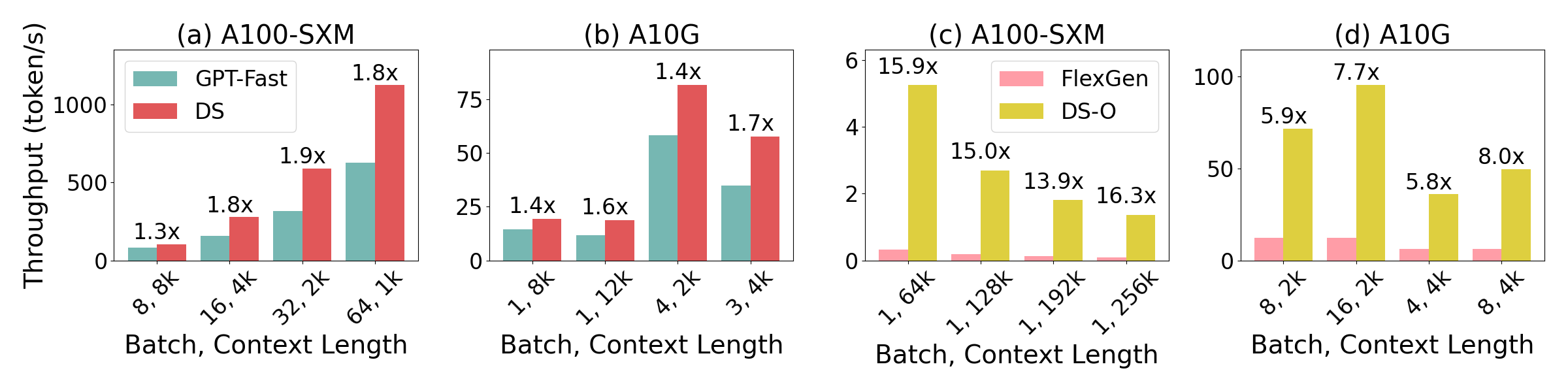} 
\caption{Throughput (token/s) and speedup of Double Sparsity (Offloading) in end-to-end scenarios.} 
\label{fig:e2e-speedup} 
\end{figure}

Figure \ref{fig:e2e-speedup} (a)(b) presents the throughput comparison between Double Sparsity and gpt-fast, measured in tokens per second across various batch sizes and sequence lengths.
We deployed the Llama-2-7B model and maximized memory usage to achieve high workload conditions.
The results indicate that Double Sparsity yields a minimum speedup of 1.3x across all tested conditions. In certain scenarios, the speedup approached twofold, showcasing Double Sparsity's overall efficiency.

In Figure \ref{fig:e2e-speedup} (c)(d), we compare the throughput of Double Sparsity-Offload to FlexGen under a constrained memory footprint, set at 1/16 of a full KV cache for both methods. Both techniques utilize a double buffer for asynchronous data copying. 
The results show that Double Sparsity-Offload achieves a 4-8$\times$ speedup over FlexGen under regular workloads, and a 16$\times$ speedup in scenarios with long texts ranging from 64K to 256K in sequence length.

\section{Related Work}
\label{sec:related_work}

\paragraph{Sparse Attention Inference}

Due to the significant overhead of attention mechanisms, many studies have focused on exploiting attention sparsity to accelerate inference.
These efforts can be categorized under three main criteria: 1) static or dynamic sparse patterns; 2) the presence of token eviction; 3) accelerating pre-filling or decoding.
StreamingLLM~\citep{xiao2024efficient} and LM-Infinite~\citep{han2023lm} utilize static sparse patterns with token eviction to accelerate decoding. These approaches achieve inference acceleration by preserving only a small number of initial tokens along with local tokens.
H2O~\citep{zhang2024h2o} and Scissorhands~\citep{liu2024scissorhands} employ dynamic sparse patterns with token eviction for decoding, preserving only a small fraction of the KV cache called heavy hitters according to accumulated attention scores, while FastGen~\citep{ge2024model} uses adaptive sparse attention patterns for different attention heads.
MInference~\citep{jiang2024minference10acceleratingprefilling} serves as a prefilling acceleration method that retains all tokens. It first identifies sparse patterns within the model, and then leverages these identified patterns to compute the pre-filling stage.
SparQ~\citep{ribar2023sparq} and Quest~\citep{tang2024quest} implement dynamic sparse decoding while also preserving all tokens. SparQ filters the top-k tokens using heavy channels of queries. Quest segments tokens into multiple pages and computes attention on the top-k pages to facilitate the decoding process.


\paragraph{Sparse Attention Training}
There are also many efforts to reduce attention complexity through training~\citep{qiu2020blockwise, ding2023longnet,tay2020sparse,chen2021scatterbrain}.
For example, 
Sparse transformer~\citep{child2019sparsetransformer} reduces the complexity to $O(n\sqrt{n})$ by introducing sparse factorization of the attention matrix.
Reformer~\citep{kitaev2019reformer} achieves $O(n\log{n})$ complexity via locality-sensitive hashing.
Longformer~\citep{beltagy2020longformer}, BigBard~\citep{zaheer2020big}, and Linformer~\citep{wang2020linformer} further reduce the complexity to linear.
Linear attention architectures have also been proposed in~\cite{katharopoulos2020transformers}.



\paragraph{Other Attention and Inference Optimizations}
Despite efforts to sparsify the attention computation, there are many other optimizations for attention efficiency.
Common techniques include quantization and compression~\citep{hooper2024kvquant,liu2024kivi,kang2024gear,nawrot2024dynamic}, efficient attention architecture like multi-query attention~\citep{shazeer2019fast} and group-query attention~\citep{ainslie2023gqa}, and memory-efficient attention algorithms~\citep{rabe2021self, dao2022flashattention}.
Alternatives to transformers include using the state space model to remove the attention mechanism~\citep{gu2021efficiently}.
Other common inference optimizations for LLMs include batching~\cite{yu2022orca}, memory optimizations~\cite{sheng2023flexgen,kwon2023vllm,aminabadi2022deepspeed}, parameter sharing~\cite{sheng2023slora,chen2023punica}, speculative decoding~\cite{stern2018blockwise, leviathan2023fast, miao2023specinfer}, scheduling~\cite{han2022microsecond,agrawal2023sarathi,patel2023splitwise,zhong2024distserve},
quantization~\cite{xiao2023smoothquant, lin2023awq, dettmers2022gpt3, frantar2022optq}, and sparsification~\cite{frantar2023sparsegpt}.


\section{Future Directions and Conclusion}

\label{sec:conclusion}

\textbf{Future Directions.}
Despite the progress made with Double Sparsity, several limitations remain that reveal promising directions for future research. It is challenging to perfectly overlap communication with computation.
Enhancing asynchronous capabilities to mask communication overheads presents a promising direction that allows for significant acceleration with a minimal memory footprint.

\textbf{Conclusion.}
In this work, we introduced Double Sparsity and Double Sparsity-Offload, innovative post-training sparse attention techniques.
Double Sparsity leverages offline calibration and label cache to achieve nearly lossless performance across various benchmarks at a 1/16 sparsity level. 
Performance tests showed that Double Sparsity could accelerate attention computations by up to 16$\times$ and achieve an end-to-end speedup of 1.9$\times$. 
Double Sparsity-Offload significantly reduced KV Cache memory usage to 1/16, outperforming the throughput of previous SOTA offloading techniques by 16 times. 


%

\bibliography{colm2024_conference}
\bibliographystyle{colm2024_conference}

\newpage
\appendix

\section{Offline Calibration Illustration}

\label{sec:appendix-offline}

\begin{figure}[htbp]
  \centering
  \begin{subfigure}[b]{0.45\textwidth}
    \includegraphics[width=\textwidth]{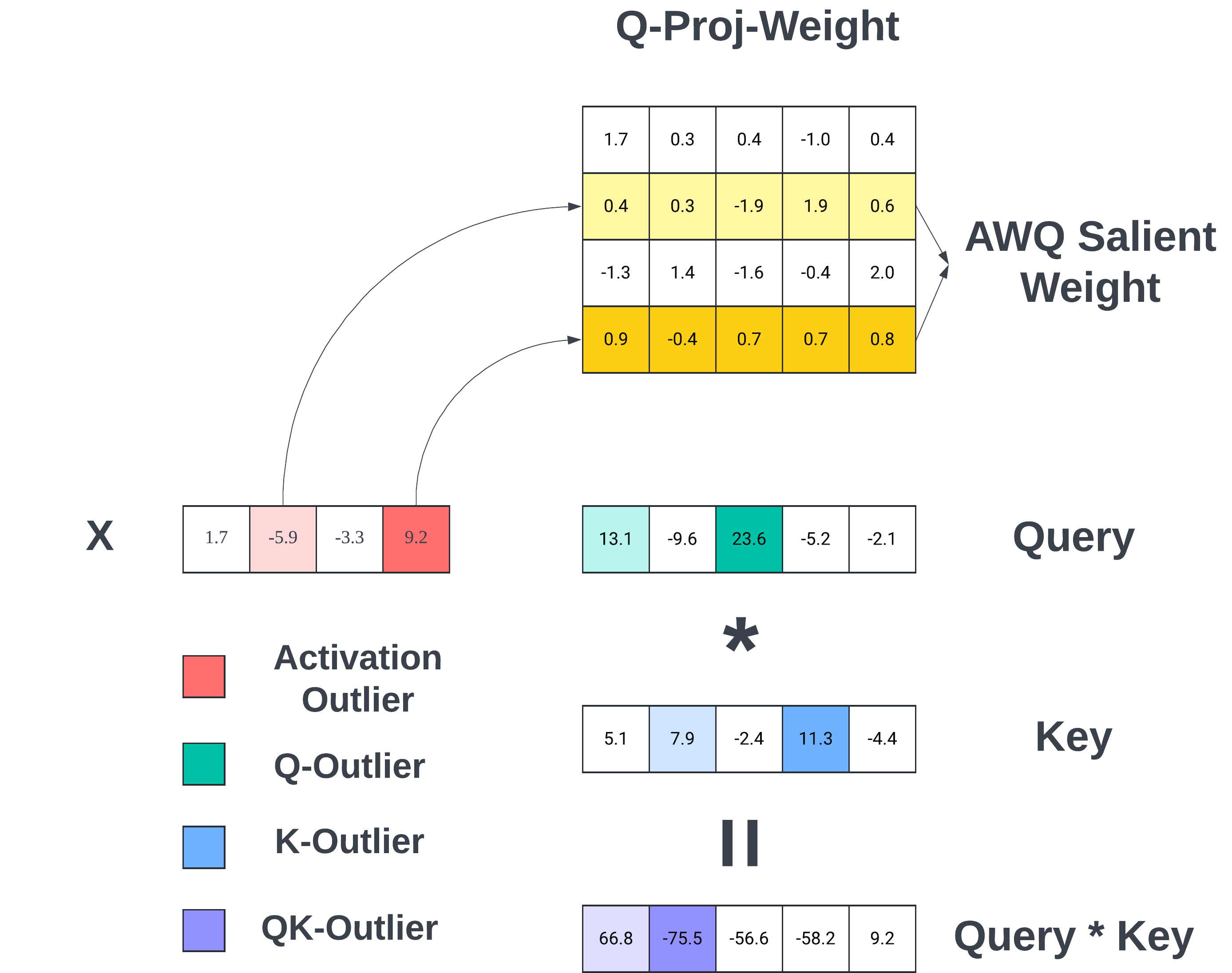}
    \caption{Outlier channels of AWQ and Double Sparsity.}
    \label{fig:offline-calibration}
  \end{subfigure}
  \hfill
  \begin{subfigure}[b]{0.45\textwidth}
    \includegraphics[width=\textwidth]{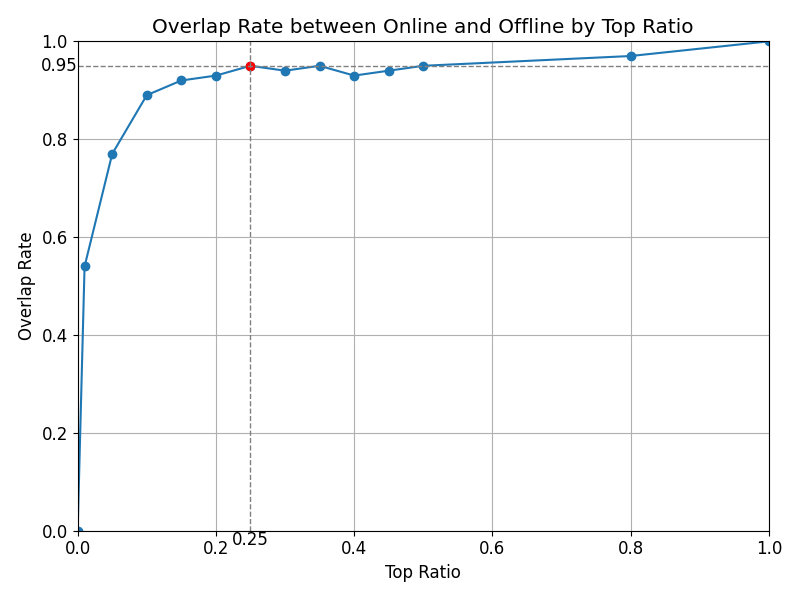}
    \caption{Outlier channel overlap rate between offline calibration and online decoding.}
    \label{fig:calibration-overlap}
  \end{subfigure}
  \caption{Analysis of Double Sparsity calibration in identifying outlier channels}
  \label{fig:test}
\end{figure}

The x-axis of the Figure \ref{fig:calibration-overlap} denotes the ratio of the selected top-k channels to the total number of channels, while the y-axis quantifies the degree of overlap between the offline and online outliers. 

\section{Ablation Study}

\subsection{Forward without Label Cache}

\label{sec:no-label-cache}

To investigate the significance of the label cache in the forward pass of Double Sparsity, we conducted experiments comparing performance with and without the label cache.
As depicted in Table \ref{tab:ablation-label}, label caches significantly enhance the forward pass, yielding a speedup ranging from 2 to 4 times.

\begin{table}[ht]
\centering
\caption{Latency comparing performance With and Without Label Cache. }
\label{tab:ablation-label}
\begin{tabular}{cccccc}
\toprule
Batch & Seq Len & With Label Cache (ms) & Without Label Cache (ms) & Speedup \\
\midrule
4 & 2048 & 0.165 & 0.279 & 1.7 \\
4 & 4096 & 0.181 & 0.559 & 3.1 \\
4 & 8192 & 0.504 & 1.250 & 2.5 \\
4 & 16384 & 1.550 & 3.000 & 1.9 \\
32 & 2048 & 0.467 & 1.960 & 4.2 \\
32 & 4096 & 0.983 & 3.950 & 4.0 \\
32 & 8192 & 3.600 & 9.540 & 2.6 \\
32 & 16384 & 12.600 & 24.000 & 1.9 \\
\bottomrule
\end{tabular}
\end{table}

\subsection{Embedding Similarity Across Layers}

\label{sec:cosine-similarity}

\begin{figure}[ht]
\centering 
\vspace{-1em}
\includegraphics[width=1\textwidth]{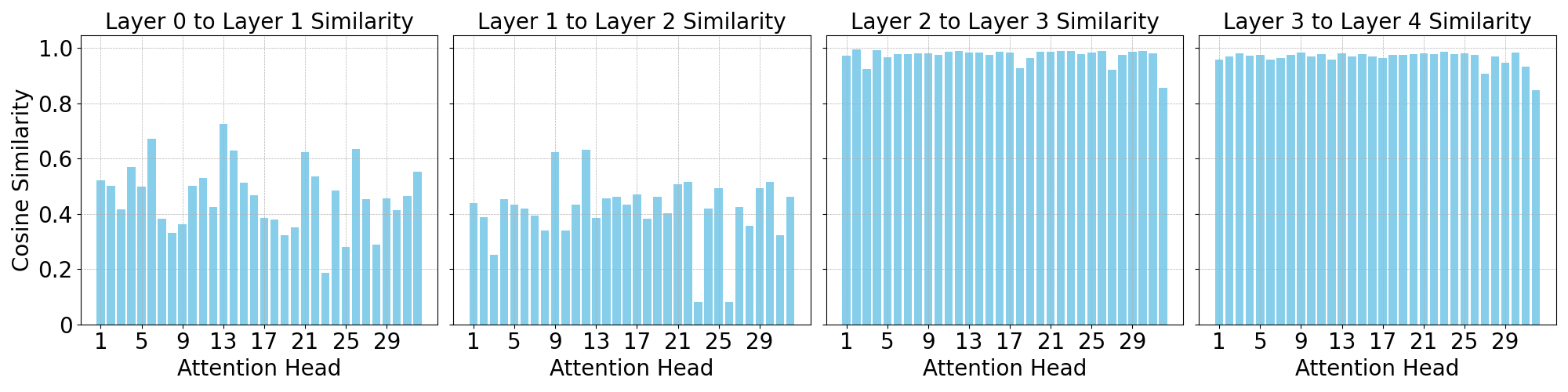} 
\caption{Average cosine similarity of embeddings across all attention heads between layers 0-1, 1-2, 2-3, and 3-4 on the Pile dataset for Llama-2-7B model.} 
\label{fig:cosine-simliarity} 
\end{figure}

\section{Perplexity Selection Illustration}

\label{sec:perplexity-select}

\begin{figure}[ht]
\centering 
\includegraphics[width=1\textwidth]{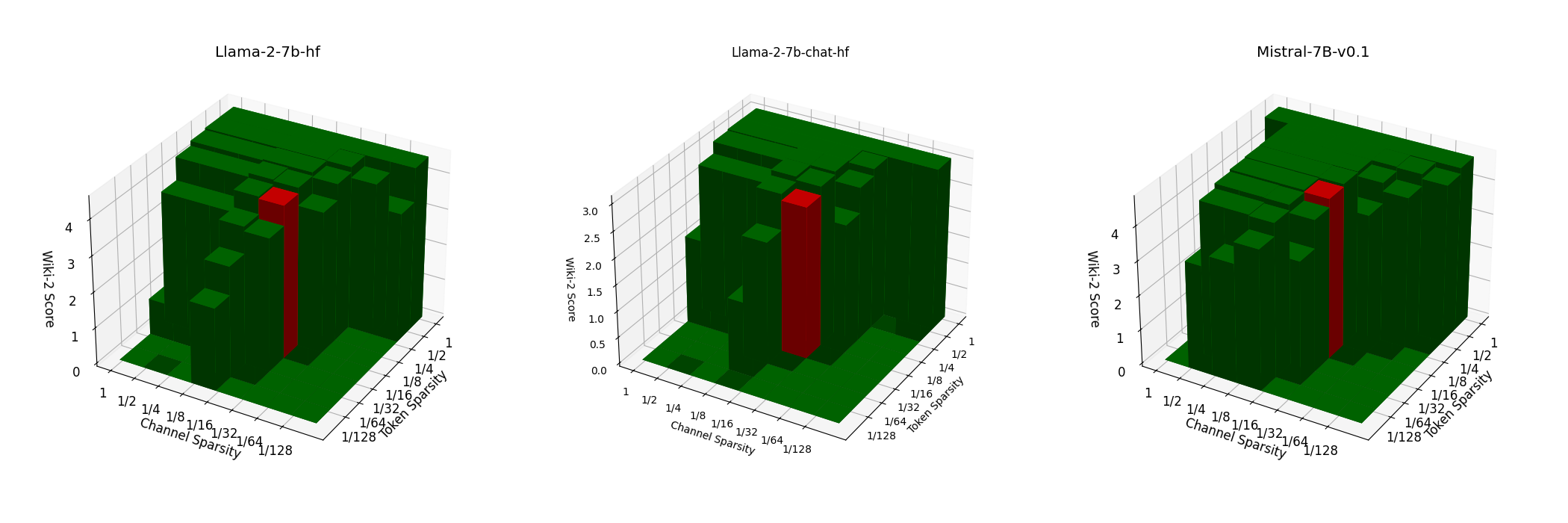}
\vspace{-1em}
\caption{Perplexity of models at different token-sparsity and channel-sparsity levels. Notably, the red bars, representing a sparsity level of 1/16 for both token and channel, show that the model's performance remains largely consistent with the original model at this level.} 
\vspace{-1em}

\label{fig:3d-bar} 
\end{figure}

Figure \ref{fig:3d-bar} uses token-level sparsity as the x-axis, channel-level sparsity as the y-axis, and 10-perplexity values as the z-axis, where higher bars indicate better performance. A sudden shift in perplexity is observed as the sparsity level goes beyond 1/16.

\newpage
\section*{NeurIPS Paper Checklist}

\begin{enumerate}

\item {\bf Claims}
    \item[] Question: Do the main claims made in the abstract and introduction accurately reflect the paper's contributions and scope?
    \item[] Answer: \answerYes{} 
    \item[] Justification: The claims in the abstract and introductions are empirically verified by the experimental section \ref{sec:exp}.

\item {\bf Limitations}
    \item[] Question: Does the paper discuss the limitations of the work performed by the authors?
    \item[] Answer: \answerYes{} 
    \item[] Justification: The limitations and future directions to address them are discussed in \ref{sec:conclusion}.

\item {\bf Theory Assumptions and Proofs}
    \item[] Question: For each theoretical result, does the paper provide the full set of assumptions and a complete (and correct) proof?
    \item[] Answer: \answerNA{} 
    \item[] Justification: We don't include theoretical results. All the results are derived from experiments.

    \item {\bf Experimental Result Reproducibility}
    \item[] Question: Does the paper fully disclose all the information needed to reproduce the main experimental results of the paper to the extent that it affects the main claims and/or conclusions of the paper (regardless of whether the code and data are provided or not)?
    \item[] Answer: \answerYes{} 
    \item[] Justification: The experimental setups are described in Section \ref{sec:accuracy-eval} and Section \ref{sec:speedup-eval}. The code is available.

\item {\bf Open access to data and code}
    \item[] Question: Does the paper provide open access to the data and code, with sufficient instructions to faithfully reproduce the main experimental results, as described in supplemental material?
    \item[] Answer: \answerYes{} 
    \item[] Justification: The code is available.

\item {\bf Experimental Setting/Details}
    \item[] Question: Does the paper specify all the training and test details (e.g., data splits, hyperparameters, how they were chosen, type of optimizer, etc.) necessary to understand the results?
    \item[] Answer: \answerYes{} 
    \item[] Justification: The experimental setups are described in Section \ref{sec:accuracy-eval} and Section \ref{sec:speedup-eval}. The code is available. The code is available.

\item {\bf Experiment Statistical Significance}
    \item[] Question: Does the paper report error bars suitably and correctly defined or other appropriate information about the statistical significance of the experiments?
    \item[] Answer: \answerYes{} 
    \item[] Justification: We measure the throughput and latency by running a sufficiently large number of tests and reporting the average. These results are highly deterministic, so we do not include error bars in the figures to maintain clarity. The improvements are statistically significant.

\item {\bf Experiments Compute Resources}
    \item[] Question: For each experiment, does the paper provide sufficient information on the computer resources (type of compute workers, memory, time of execution) needed to reproduce the experiments?
    \item[] Answer: \answerYes{} 
    \item[] Justification: The experimental setups are described in Section \ref{sec:speedup-eval}. We use common NVIDIA GPUs to conduct experiments. The code is available.
    
\item {\bf Code Of Ethics}
    \item[] Question: Does the research conducted in the paper conform, in every respect, with the NeurIPS Code of Ethics \url{https://neurips.cc/public/EthicsGuidelines}?
    \item[] Answer: \answerYes{} 
    \item[] Justification: The research respects NeurIPS Code of Ethics.

\item {\bf Broader Impacts}
    \item[] Question: Does the paper discuss both potential positive societal impacts and negative societal impacts of the work performed?
    \item[] Answer: \answerNA{} 
    \item[] Justification: The techniques introduced in this paper make the use of large language models easier and more efficient. They do not have additional direct societal impact beyond the existing societal impact of large language models. However, because they accelerate the execution of LLMs, they may amplify the existing societal impact of these models, whether positive or negative.
    
\item {\bf Safeguards}
    \item[] Question: Does the paper describe safeguards that have been put in place for responsible release of data or models that have a high risk for misuse (e.g., pretrained language models, image generators, or scraped datasets)?
    \item[] Answer: \answerNA{} 
    \item[] Justification: This paper does not release new datasets or models, so this is not applicable.

\item {\bf Licenses for existing assets}
    \item[] Question: Are the creators or original owners of assets (e.g., code, data, models), used in the paper, properly credited and are the license and terms of use explicitly mentioned and properly respected?
    \item[] Answer: \answerYes{} 
    \item[] Justification: This paper uses open-weight models and public datasets for experiments. The usage respects all the original licenses.

\item {\bf New Assets}
    \item[] Question: Are new assets introduced in the paper well documented and is the documentation provided alongside the assets?
    \item[] Answer: \answerNA{} 
    \item[] Justification: This paper doesn't release new assets.

\item {\bf Crowdsourcing and Research with Human Subjects}
    \item[] Question: For crowdsourcing experiments and research with human subjects, does the paper include the full text of instructions given to participants and screenshots, if applicable, as well as details about compensation (if any)? 
    \item[] Answer: \answerNA{} 
    \item[] Justification: This paper does not involve crowdsourcing nor research with human subjects.

\item {\bf Institutional Review Board (IRB) Approvals or Equivalent for Research with Human Subjects}
    \item[] Question: Does the paper describe potential risks incurred by study participants, whether such risks were disclosed to the subjects, and whether Institutional Review Board (IRB) approvals (or an equivalent approval/review based on the requirements of your country or institution) were obtained?
    \item[] Answer: \answerNA{} 
    \item[] Justification: This paper does not involve crowdsourcing nor research with human subjects.

\end{enumerate}

\end{document}